\documentclass[10pt,twocolumn,letterpaper]{article}

\usepackage{wacv}
\usepackage{times}
\usepackage{epsfig}
\usepackage{graphicx}
\usepackage{amsmath}
\usepackage{amssymb}
\usepackage[accsupp]{axessibility}  
\usepackage{tabularx, booktabs} 
\usepackage{emp}
\usepackage{pifont}
\usepackage{empheq}
\usepackage{multirow}
\newcolumntype{P}[1]{>{\centering\arraybackslash}p{#1}}
\newcolumntype{M}[1]{>{\centering\arraybackslash}m{#1}}
\newcommand{\xmark}{\ding{55}}%
\def\wacvPaperID{1133} 

\wacvfinalcopy 

\ifwacvfinal
\fi

\ifwacvfinal
\usepackage[breaklinks=true,bookmarks=false]{hyperref}
\else
\usepackage[pagebackref=true,breaklinks=true,colorlinks,bookmarks=false]{hyperref}
\fi

\begin{document}

\title{Sign Language Translation with Hierarchical Spatio-Temporal\\ Graph Neural Network}

\author{Jichao Kan\textsuperscript{1,2,}\thanks{Work was done during Master of Philosophy Study at The University of Sydney.}, Kun Hu\textsuperscript{1,}\thanks{Corresponding Author, supported by Australian Research Council (ARC) Grant DP210102674.}
, Markus Hagenbuchner\textsuperscript{3}, Ah Chung Tsoi\textsuperscript{3}, \\Mohammed Bennamoun\textsuperscript{4}, Zhiyong Wang\textsuperscript{1}\\
\textsuperscript{1}School of Computer Science, The University of Sydney\\
\textsuperscript{2}Data Science Institute, University of Technology Sydney\\
\textsuperscript{3}School of Computing and Information Technology, University of Wollongong\\
\textsuperscript{4}Department of Computer Science and Software Engineering, The University of Western Australia\\
{\tt\small jichao.kan@student.uts.edu.au}, {\tt\small kuhu6123@uni.sydney.edu.au}, 
{\tt\small \{markus,act\}@uow.edu.au}, \\
{\tt\small mohammed.bennamoun@uwa.edu.au}, {\tt\small  zhiyong.wang@sydney.edu.au}
}

\maketitle

\ifwacvfinal
\thispagestyle{empty}
\fi

\begin{abstract}
   Sign language translation (SLT), which generates text in a spoken language from visual content in a sign language, is important to assist the hard-of-hearing community for their communications. Inspired by neural machine translation (NMT), most existing SLT studies adopted a general sequence to sequence learning strategy. However, SLT is significantly different from general NMT tasks since sign languages convey messages through multiple visual-manual aspects. Therefore, in this paper, these unique characteristics of sign languages are formulated as hierarchical spatio-temporal graph representations, including high-level and fine-level graphs of which a vertex characterizes a specified body part and an edge represents their interactions. Particularly, high-level graphs represent the patterns in the regions such as hands and face, and fine-level graphs consider the joints of hands and landmarks of facial regions. To learn these graph patterns, a novel deep learning architecture, namely hierarchical spatio-temporal graph neural network (HST-GNN), is proposed. Graph convolutions and graph self-attentions with neighborhood context are proposed to characterize both the local and the global graph properties. Experimental results on benchmark datasets demonstrated the effectiveness of the proposed method.
\end{abstract}

\section{Introduction}

Sign languages, which engage visual-manual modalities to convey meanings, are the primary communication tools for the deaf and hard-of-hearing community. However, it is still an open research problem to reduce the gap of the communications between sign language users and spoken language users who have limited sign language knowledge. Therefore, researchers have utilized various methods to convert sign language to spoken language for a better communication between the users of different languages. 



Early efforts fell into the category of sign language recognition (SLR). At the beginning, SLR methods aimed to recognize an isolated gloss from a sign language video \cite{grobel1997isolated, uebersax2011real}, while the continuous nature of languages was ignored. Therefore, continuous SLR was proposed to generate a sequence of pre-defined glosses (i.e., the written words interpreting signing poses) \cite{cooper2012sign, Koller2016}. With the recent success of deep learning techniques in many applications, SLR can be regarded as a neural machine translation (NMT) tasks following an encoder-decoder framework, where the \textit{source} is a \textit{video} and the \textit{target} is a corresponding \text{spoken sentence}. Early NMT based studies for sign language understanding conducted a SLR task (e.g., \cite{cooper2012sign, grobel1997isolated, Koller2015}). Researchers adopted an encoder to extract latent representations from sign language videos and a decoder to perform the process of generating gloss scripts \cite{Camgoz2017}. Similar to the development of general NMT studies, attention mechanisms \cite{Bahdanau2015, Vaswani2017} have been introduced for SLR to focus on the most relevant video frames when generating a specified gloss~\cite{Zhou2020}. 

It is worth noting that sign languages convey meaning from multiple aspects: the manual articulations and the non-manual elements such as postures and movements of different body parts contribute to the meaning as well as the lexical distinction, grammatical structure, adjectival or adverbial content, and discourse functions. It is anticipated that such unique domain knowledge could be beneficial for advancing sign language understanding. 
As a result, a number of studies were proposed to explore such knowledge by characterizing the local patterns individually of several key regions of a signer's body \cite{Camgoz2017,huang2018video,Ko2018,Koller2015,Liu,Sridhar2020}. Recently, the interactions between these local regions have been explored by focusing only on the spatial relations between the regions or the temporal relations across the same region \cite{Koller2016,Koller2018,Zhou2020, 9423424}. 
Nonetheless, these SLR methods are not ideal for sign language understanding as the gloss-level recognition is still different from spoken languages. 
Therefore, it is attractive to devise sign language translation (SLT) methods to reduce the communication barrier. SLT takes a sign language sentence performed by a signer as input to produce text scripts of the signing sentence in a spoken language. 
As an early attempt, \cite{Camgoz2018} formalized the SLT problem and released an SLT dataset PHONEIX-Weather-2014-T. After that, more MNT based methods were investigated for SLT tasks (e.g., \cite{guo2018hierarchical, yin2020sign}). 

However, existing methods on either SLR and SLT have not fully explored the interactions between those local regions at a fine-grained level to best utilize the unique aspects of sign languages, which demands a better representation of signing poses. 
Therefore, in this paper, to better represent the spatio-temporal relations at a finer-grained level for SLT, a novel graph-based sign language representation and a hierarchical graph neural network architecture are proposed. 
As illustrated in \autoref{fig:encoder}, a sign language sentence can be characterized with appearance, motion and pose representations. The proposed method exploits these representations from a fine-level to a high-level. A fine-level spatio-temporal graph is based on the joints within a human body region (e.g., the left-hand). 
A high-level spatio-temporal graph is to formulate the relations between the human body regions, which can be based on appearance, motion, or pose features. 
To this end, a hierarchical spatio-temporal graph neural network (HST-GNN) to learn the hierarchical graph patterns with both high-level and fine-level graphs for SLT.
HST-GNN introduces a connection between graph convolutions and graph self-attentions using the neighborhood context, which helps to formulate graph patterns from multiple perspectives. Comprehensive experiments on datasets demonstrated the performance of our proposed method. 

In summary, the key contributions of this study are as follows:
\begin{itemize}
    \item A novel deep architecture, namely hierarchical spatio-temporal graph neural network (HST-GNN), is devised for SLT.
    \item Multiple spatio-temporal graphs with hierarchical structures are constructed to represent signing poses. 
    \item Graph convolution and graph self-attention with connections based on their neighborhood context are studied to learn graph representations from multiple perspectives.  
\end{itemize}


\section{Related Work}
\label{sec:relatedwork}

In this section, relevant studies are reviewed from two aspects: existing SLR and SLT methods, and graph neural networks which are relevant to our proposed method. 

\subsection{Sign Language Recognition \& Translation}


Vision-based sing language recognition and translation aim to understand visual contents of sign languages to generate glosses and spoken language text scripts, respectively. 

Sign language recognition is a task taking the visual content performed by sign language signers to produce the associated glosses.
Early studies on SLR focused on recognizing isolated signs or gestures to produce word-level or phrase-level outputs, which followed a pattern recognition pipeline: various hand-crafted visual features such as SIFT and SURF were obtained from an input signing video and a trained classifier took these features to produce signing labels \cite{grobel1997isolated, uebersax2011real}. 
With the advances of deep learning, convolution neural networks (CNN) and recurrent neural networks (RNN) were also adopted for isolated SLR \cite{zhu2018continuous}. 
However, recognizing isolated signs provides limited understanding of a complete sign language sentence. 
Therefore, continuous SLR has been investigated at the sentence level by treating continuous signings as a sequence of signing poses. 
Recently, 
various deep architectures have been proposed to perform continuous SLR task~\cite{Guo2019, guo2018hierarchical}. To consider the domain knowledge of sign languages, fine-level regional patterns were investigated for accurate SLR in an independent manner~\cite{Camgoz2018,cui2019deep,Koller2015, pu2020boosting,pu2019iterative,yin2020sign}. 
To explore the interactions between the local regions of a human body in a signing pose, graph-based neural networks were proposed to formulate the spatial relations between the regions or the temporal relations within a region across frames~\cite{Ko2018,Koller2016, Koller2018, Li2019,Zhou2020}. 
However, the spatio-temporal relations have been seldom explored with hierarchical structures, which could miss important sign language patterns.

Although there have been impressive SLR results, the gap between glosses and spoken language sentences still exist. To address this problem, sign language translation has been studied to take the rich grammatical structures in spoken languages into consideration. It aims to generate spoken language sentences rather than a sequence of glosses~\cite{10.1145/3394171.3413931}. 
For example, RNN~\cite{Camgoz2018} and Hierarchical LSTM \cite{Ko2018} were adopted to extract visual information and to generate spoken language sentences. Moreover, the above-mentioned deep learning based SLR methods were also explored for SLT~\cite{yin2020sign, yin-read-2020-better,Zhou2020}. However, domain knowledge of sign languages such as the interactions between human body regions has not been adequately investigated yet. Missing such fine-grained patterns could result in less accurate sign language representations, and thus negatively impacts the quality of the generated spoken language scripts. 

\subsection{Graph Neural Network}

While conventional neural networks have been utilized to process vectorized data, there are numerous applications involving data in non-Euclidean forms such as graphs. 
Graph neural network (GNN) was first proposed to address the learning tasks with graph inputs \cite{hu2021graph,hu2019graph,scarselli2009graph,wu2020comprehensive}. Graph convolution network (GCN) extended convolution filters for graphs to help construct deep graph representations~\cite{bruna2013spectral}. Graph attention network (GAT) was further proposed to estimate adjacency weights \cite{velickovic2018graph}. 
Recently, various methods have been proposed to discover special graph properties, which have been ignored by conventional GCNs, such as CPNGNN \cite{sato2019approximation} which exploits the local ordering of the nodes to increase the representation capacity of graph networks, and DimNet~\cite{Klicpera2020Directional} which introduces directed message passing to improve the capacity of graph representation. \cite{bull2020automatic} applied the GCN in the sign language segmentation.

\section{Methodology}

\label{sec:method}
\begin{figure*}
    \centering
    \includegraphics[width=\textwidth]{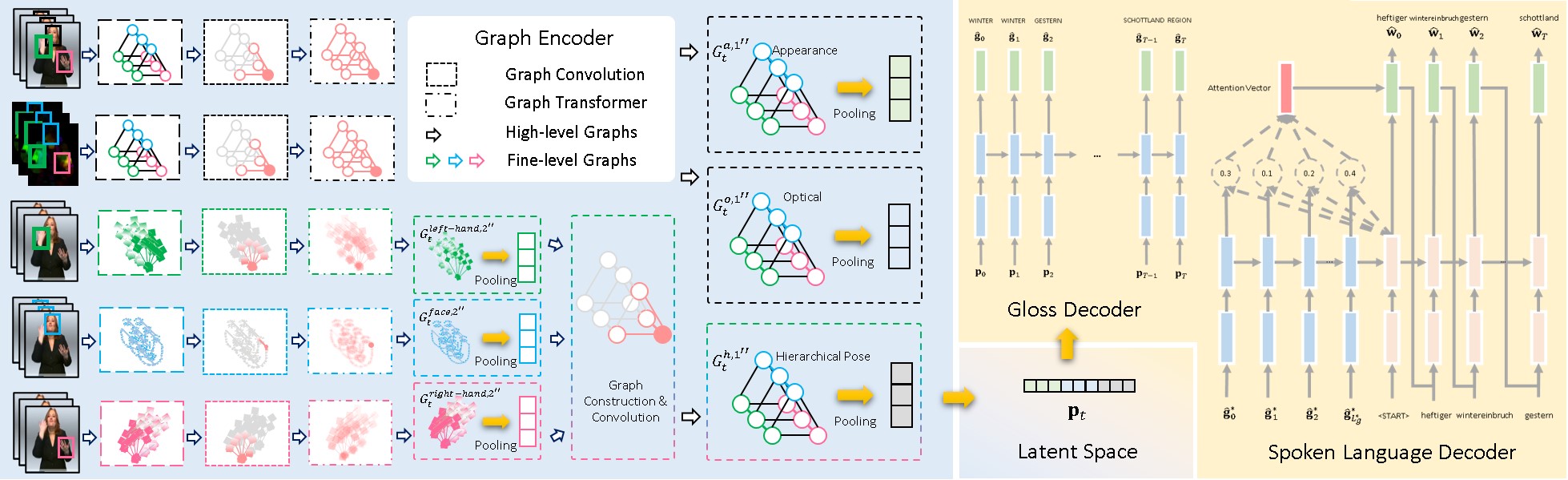}
    \caption{Illustration of the proposed HST-GNN architecture for SLT. 
    }
    \label{fig:encoder}
\end{figure*}

The proposed HST-GNN adopts an encoder-decoder scheme as illustrated in Figure~\ref{fig:encoder}. The encoder represents an input video of a sign language sentence to a latent space, which includes graph construction, graph convolution and graph self-attention mechanism. Two levels of graphs are constructed including high-level graphs between the key body regions (i.e., the left and right hands and the face) with appearance and optical flow vertex features and fine-level graphs of the key regions with appearance features (i.e., left-hand graph, right-hand graph and facial graph). The graph convolution and the graph self-attention with neighborhood context are introduced to formulate the local and the global graph properties, respectively. A hierarchical graph pooling mechanism is adopted to fuse these graphs as the final encoded latent vector for an input sign language sentence.

The decoder generates text scripts in a spoken language using the encoded representation, which adopts
a two stage recognition scheme with two LSTMs containing attention mechanisms: the first one translates the fused vector to glosses (i.e., written words corresponding to individual signing poses) and the second one translates the glosses to text scripts in a spoken language. 
In this section, the details of the proposed methods are introduced. 

\subsection{Hierarchical Spatio-temporal Graphs}
As illustrated in Figure~\ref{fig:encoder}, two levels of graphs including high-level graphs and fine-level graphs are constructed to characterize the three key-regions of a signer's body and their interactions in hierarchical structures. 

A high-level graph characterizes the spatio-temporal relationships between the three key regions of a human body in video frames.
That is, a high-level graph can be constructed with three vertices which denote facial region, left-hand region, and right-hand region, respectively. 
The three key regions can be obtained by pose estimation algorithms (e.g., HRNet~\cite{wang2020deep}). To characterize each vertex, a bounding box is used to extract an associated frame sub-patch and the vertex-level feature can be computed using the image patch with a pre-trained CNN. Note that the frame used to extract the visual features can be in the modality of appearance (RGB) or motion (e.g., optical flow\cite{zach2007duality}) in this paper. As a result, two high-level graphs are constructed, one for each modality. 

In more detail, denote the vertex-level features for high-level graphs (Level 1 graphs) derived from the $t$-th frame of the input video as $\dot{\mathbf{V}}_{t}^{m,1} = \{\dot{\mathbf{v}}_{i,t}^{m,1} \in \mathbb{R}^{d}\}$, where $d$ is the dimension of the feature vectors of modality $m\in \{{\rm appearance (a), optical flow (o)}\}$ and $i$ indicates the $i$-th body region in the frame. 
To involve additional temporal relationship from neighbouring frames, a slide window $W$ is defined and the set of the vertices of the $t$-th graph is defined as follows:   
\begin{equation}
\mathbf{V}_t^{m,1} = \bigcup\limits_{w=-W}^W \dot{\mathbf{V}}_{t + w}^{m,1} := \{\mathbf{v}_{i,t}^{m,1} \in \mathbb{R}^{d}|i=1,...,n_{1}\}, 
\end{equation}
which contains $n_{1}$ vertices.

An adjacency matrix $\mathbf{A}_{t}^{m,1}$ is defined to represent the relationships between the vertices in $\mathbf{V}_{t}^{m,1}$. 
Instead of using a pre-defined $\mathbf{A}_{t}^{m,1}$ empirically, in this study, $\mathbf{A}_{t}^{m,1}$ is learned in an unsupervised way. The element $a_{ij,t}^{m,1}$ of $\mathbf{A}_{t}^{m,1}$ in the $i$-th row and $j$-th column denotes the interaction between the vertices $\mathbf{v}_{i,t}^{m,1}$ and $\mathbf{v}_{j,t}^{m,1}$. 
The following computation characterizes the interaction $a_{ij,t}^{m,1}$ by considering the relevant vertex-level features:
\begin{equation}
    a_{ij,t}^{m,1} = \sigma({\mathbf{v}_{i,t}^{m,1}}^{T}\mathbf{M}^{m,1}\mathbf{v}_{j,t}^{m,1}),
\end{equation}
where $\mathbf{M}^{m,1}\in \mathbb{R}^{d \times d}$ is the matrix of a bilinear transform containing trainable parameters and $\sigma$ is an non-linear function to increase the capability to represent complex patterns. 
Furthermore, to reduce the number of the parameters in $\mathbf{M}^{m,1}$ and the model complexity, a low-rank decomposition of $\mathbf{M}^{m,1}$ is introduced:
\begin{equation}
    \mathbf{M}^{m,1} = \mathbf{M}^{m,1}_{1}{\mathbf{M}^{m,1}_{2}}^{T},
\end{equation}
where $\mathbf{M}^{m,1}_{1}\in\mathbb{R}^{d\times p}$, $\mathbf{M}^{m,1}_{2}\in\mathbb{R}^{d\times p}$, and $p<<d$.

In this paper, two additional constraints are applied to the adjacency matrix $\mathbf{A}^{m,1}_{t}$. The first one is for symmetry:
\begin{equation}
    \dot{\mathbf{A}}_{t}^{m,1} = {\mathbf{A}_{t}^{m,1}}^{T}\mathbf{A}_{t}^{m,1}.
\end{equation}
The second one is for the normalization, which alleviates the scale difference between the vertices. In detail, $\dot{\mathbf{A}}$ is normalized by its matrix norm: 
\begin{equation}
    \ddot{\mathbf{A}}_{t}^{m,1} = \frac{\dot{\mathbf{A}}_{t}^{m,p}}{||\dot{\mathbf{A}}_{t}^{m,1}||}.
\end{equation}
For convenience, in the following discussion, $\mathbf{A}_{t}^{m,1}$, $\dot{\mathbf{A}}_{t}^{m,1}$ and $\ddot{\mathbf{A}}_{t}^{m,1}$ are not particularly distinguished. 
In summary, for a particular modality $m$, a high-level graph sequence can be obtained as 
$\{\mathbf{G}^{m,1}_{t}=\{\mathbf{V}^{m,1}_{t}, \mathbf{A}^{m,1}_{t}\}\}$. 

For the fine-level graphs, key points (e.g., the joints for hands and the landmarks for face) are further identified to construct individual graphs (Level 2 graphs) for each body region (i.e., left hand, right hand, and face). Similar to the high-level graphs, the regions near the joints can be used to characterize these key points and their relationships. For computational efficiency, only appearance features are used for the fine-level graphs and $\{\mathbf{G}^{r,2}_{t}=\{\mathbf{V}^{r,2}_{t}, \mathbf{A}^{r,2}_{t}\}\}$ can be obtained, 
where $r\in\{$left hand, right hand, face$\}$ indicates the body regions under consideration.

The proposed encoder of HST-GNN encodes these graphs to a latent graph space for SLR by considering the local and the global graph properties. For convenience, the superscripts are omitted in the following discussion of the graph convolution and the graph self-attention. 


\subsection{Graph Convolution}

Graph convolution neural networks generalize the convolution filters for graph inputs of arbitrary structures, of which the input and the output are graphs. 
In detail, it can be viewed as the message passing through the neighbors of each vertices in a graph in line with its vertex-level features and adjacency patterns. By stacking multiple graph convolution layers, a deep graph neural network can be obtained for graph-based deep representations. 
Formally, the computations for the $t$-th graph of a video in the $l$-th graph convolution layer can be formulated as:
\begin{equation}
    \mathbf{H}^{l+1}_{t}=f(\mathbf{H}_{t}^{l}, \mathbf{A}_{t}; \mathbf{W}^{l}) = f(\mathbf{A}_{t}\mathbf{H}^{l}_{t}\mathbf{W}^l),
\end{equation}
where $\mathbf{H}^{l}_{t} \in \mathbb{R}^{p^l}$ is the input of the layer, $\mathbf{H}^{l+1}_{t}\in\mathbb{R}^{p^{l+1}}$ is the output of the layer, $\mathbf{W}^l\in\mathbb{R}^{p^l\times p^{l+1}}$ contains the learnable parameters as a linear transform, and $f$ is a non-linear vertex-wise activation function. 
Note that the multiplication of $\mathbf{A}_{t}$ and $\mathbf{H}^{l}_{t}$ implements the message passing through the vertices, which helps each vertex to obtain proper patterns from its neighbouring body regions or key points for aggregating and modeling the graph context. 
In this way, the graph convolutions mainly focus on the local properties for each vertex. 
In particular, $\mathbf{H}^{0}_{t}=[\mathbf{v}_1^t, \mathbf{v}_2^t, ..., \mathbf{v}_n^t]^T$ and $p^0$ is the dimension of the vertex-level feature vectors.
The output graph from the last graph convolution layer is denoted as $\mathbf{G}^\prime_{t}=\{\mathbf{V}^\prime_{t}, \mathbf{A}^\prime_{t}\}$. 

\subsection{Graph Transformer}

The transformer with the self-attention mechanism and the positional encoding was proposed for sequentially organized data \cite{Vaswani2017}, which processes the sequential patterns globally in a parallel manner. 
As graph convolutions mainly formulate the local graph properties, inspired by the transformer, we propose a novel graph transformer to formulate the global graph properties with a shortcut connection of the graph convolutions by involving the neighboring patterns. 

The self-attention mechanism can be regarded as a querying process with queries, keys and values. For an input graph $\mathbf{G}^\prime_{t}=\{\mathbf{V}^\prime_{t}, \mathbf{A}^\prime_{t}\}$, the computations are as follows: 
\begin{equation}
    \mathbf{Q}_{t} = \mathbf{V}^\prime_{t} \mathbf{W}^{Query},  \mathbf{K}_{t} = \mathbf{V}^\prime_{t} \mathbf{W}^{Key},
    \mathbf{L}_{t} = \mathbf{V}^\prime_{t} \mathbf{W}^{Value},
\end{equation}
where $\mathbf{W}^{Query}$, $\mathbf{W}^{Key}$ and $\mathbf{W}^{Value}$ are learnable parameters of three linear projections. 
With these computations, for a particular vertex ${\mathbf{v}_i^{t}}^{\prime} \in \mathbf{V}_{t}^\prime$, the associated query $\mathbf{q}_i^t$, which is the $i$-th row of $\mathbf{Q}_{t}$, is used to query a context from other vertices. More specifically, the inner product of $\mathbf{q}_i^{t}$ and $\mathbf{k}_j^{t}$ (the $j$-th row of $\mathbf{K}_{t}$) is computed as a score $s_{ij}$ to measure the extent of collecting patterns from the values $\mathbf{l}_{j}^{t}$ (the $j$-th row of $\mathbf{L}_{t}$) of the vertex ${\mathbf{v}_j^{t}}^{\prime}$ to the vertex ${\mathbf{v}_i^{t}}^{\prime}$. 
Formally, the score matrix computations can be written as:
\begin{equation}
    \mathbf{S}_{t}=\mathbf{Q}_{t}{\mathbf{K}_{t}}^{T}.
\end{equation}
To eliminate the effects in terms of the variable graph nodes, a normalization step is introduced:
\begin{equation}
    \dot{\mathbf{S}}_{t} = softmax(\frac{\mathbf{S}_{t}}{\sqrt{n}}),
\end{equation}
where $n$ is the number of the vertices of an input graph. In order to incorporate the guidance of the local graph properties, a shortcut for the neighborhood context is established with the estimated adjacency matrix $\mathbf{A}_{t}$ that is used for graph convolutions. The score matrix is computed as follows: 
\begin{equation}
    \ddot{s}_{ij} = \frac{a_{ij}exp(\dot{s_{ij}})}{\sum_ka_{ik}exp(\dot{s}_{ik})}.
\end{equation}
For the sake of convenience, $\mathbf{S}_{t}$, $\dot{\mathbf{S}}_{t}$ and $\ddot{\mathbf{S}}_{t}$ are all represented by $\mathbf{S}_{t}$ in the following discussion. 
With the score matrix, the vertex-level features with contextual information can be computed as:
\begin{equation}
    \dot{\mathbf{V}}^{\prime\prime}_{t}=\mathbf{S}_{t}\mathbf{L}_{t}.
\end{equation}
Therefore, with the above computations, the self-attention mechanism helps each vertex to collect information globally compared with graph convolutions which focus only on local neighborhoods.

In addition, to take different perspectives for the graph modelling, multiple independent self-attention heads can be computed as $\mathbf{V}_{t, k}''$, where $k \in \{1, ..., K_{head}\}$ and $K_{head}$ is the number of the independent heads. 
A fully connected layer can be used to summarize these concatenated attention heads as: 
\begin{equation}
    \dot{\mathbf{V}}^{\prime\prime}_{t} = [\dot{\mathbf{V}}_{t, 1}'', \dot{\mathbf{V}}_{t, 2}'', ... , \dot{\mathbf{V}}_{t, n}'']\mathbf{W}^{Multihead},
\end{equation}
where $\mathbf{W}^{Multihead}$ is the parameter matrix of a linear projection.

Lastly, a vertex-level feed-forward network (FFN) is applied to $\dot{\mathbf{V}}^{\prime\prime}_{t}$, which is constructed by fully connected layers. 
The computations in the transformer are computed in a vertex-wise manner, so the outputs are still in graphs. 
In addition, the graph transformer can also be stacked to construct deep architectures: each of them takes the inputs from the outputs of its previous layer. The output of the graph transformer is denoted as $\mathbf{G}^{''}_{t}=\{\mathbf{V}^{''}_{t}, \mathbf{A}^{''}_{t}\}$. 

\subsection{Hierarchical Graph Pooling}

With the above discussions, the graphs at the two levels, $\mathbf{G}^{m,1}_{t}$ and $\mathbf{G}^{r,2}_{t}$, have been encoded to a latent space by using graph convolutions and graph transformers. Particularly, in the latent space, $\mathbf{G}_{t}^{a,1\prime\prime}$, $\mathbf{G}_{t}^{o,1\prime\prime}$, $\mathbf{G}_{t}^{left-hand,2\prime\prime}$, $\mathbf{G}_{t}^{right-hand,2\prime\prime}$ and $\mathbf{G}_{t}^{face,2\prime\prime}$ can be obtained as encoded graph representations for high-level appearance, high-level motion, fine-level left hand, fine-level right hand and fine-level face, respectively. To use these graphs for decoding, a hierarchical pooling strategy is introduced as illustrated in Figure~\ref{fig:encoder}. First, average pooling is applied to each fine-level graph individually and a pooled feature vector can be obtained. These feature vectors are further used to construct another high-level graph denoted as $\mathbf{G}_{t}^{h,1\prime\prime}$. Next, the high-level graphs are pooled individually and the vectors obtained can be concatenated as a fused latent vector $\mathbf{p}_{t}$ to represent the $t$-th video frame and a sequence $\mathbf{p}=\{\mathbf{p}_{0},...,\mathbf{p}_{T}\}$ is used to represent the entire video sequence. 

\subsection{Language Decoder}

Following a two-stage scheme, the language decoder aims to generate a translation in a spoken language by using the fused latent vector. 
Based on the latent vector sequences $\mathbf{p}$, the first stage - \textit{feats2gloss} - outputs an estimated gloss $\hat{\mathbf{g}}_{i}$ for each video frame (i.e., latent vector) and a sequence can be obtained as $\hat{\mathbf{g}} = \{\hat{\mathbf{g}}_{0},...,\hat{\mathbf{g}}_{T}\}$. 
The sequence $\hat{\mathbf{g}}$ can be viewed as an alignment path to video frames of an estimated gloss sequence $\hat{\mathbf{g}}^{*} = \{\hat{\mathbf{g}}^{*}_{0},...,\hat{\mathbf{g}}^{*}_{L_g^{*}}\}$ of the ground truth $\mathbf{g}^{*}=\{\mathbf{g}_{0}^{*},...,\mathbf{g}_{L_{g^{*}}}^{*}\}$, where $L_{g^{*}}$ is the length of the gloss sequence. 
The second stage - \textit{gloss2text} - outputs the sentence $\hat{\mathbf{w}} = \{\hat{\mathbf{w}}_{0},...,\hat{\mathbf{w}}_{L_w}\}$ as an estimation of the ground truth $\mathbf{w} = \{\mathbf{w}_{0},...,\mathbf{w}_{L_w}\}$ in spoken language using the estimated gloss sequence $\hat{\mathbf{g}}^{*}$, where ${L_w}$ is the sentence length.

The \textit{feats2gloss} stage involves an LSTM network for sequential to sequential recognition. Formally, this stage formulates the following conditional probability:
\begin{equation}
    p(\mathbf{g}_{t}|\hat{\mathbf{g}}_{0},...,\hat{\mathbf{g}}_{t-1},\mathbf{p}_{0},...,\mathbf{p}_{t-1}),
\end{equation}
which is the probability that the $t$-th generated gloss is $\mathbf{g}_t$ by considering the previously generated glosses $\hat{\mathbf{g}}_{0},...,\hat{\mathbf{g}}_{t-1}$ and the encoded vectors $\mathbf{p}_{0},...,\mathbf{p}_{t-1}$. 
In particular, denote the output of this LSTM network as $\mathbf{Y}^{g} = (y^{g}_{ij})$, in which the element $y^{g}_{ij}$ indicates the probability that the $i$-th gloss in the output sequence is associated with the $j$-th encoded latent vector.

The \textit{gloss2text} stage is based on another LSTM network with a general attention mechanism. It adopts the generated gloss sequence to formulate the following probability: 
\begin{equation}
    p(\mathbf{w}_{l}|\hat{\mathbf{w}}_{0},...,\hat{\mathbf{w}}_{l-1},\hat{\mathbf{g}}_{0}^{*},...,\hat{\mathbf{g}}_{L_{g}^{*}}),
\end{equation}
where the estimation $\hat{\mathbf{w}}_l$ of the $l$-th word $\mathbf{w}_l$ is obtained according to the previously estimated words $\hat{\mathbf{w}}_{0},...,\hat{\mathbf{w}}_{l-1}$ and the gloss sequence $\hat{\mathbf{g}}^{*}$ estimated by decoding stage 1. The generation starts from $\mathbf{w}_{start}$, which is a start signal, and ends with a stopping signal $\mathbf{w}_{end}$. 

\subsection{Optimization Loss}

The translation error is measured by considering the error of the generated glosses and the generated words, associated with the two decoding stages - \textit{feats2gloss} and \textit{gloss2text}, respectively. 

Following a conventional practice in SLR, a connectionist temporal classification (CTC) loss \cite{Graves2006} is adopted for the glosses, which helps to obtain the unknown alignment between the encoded vector sequence and the gloss sequence. 
In detail, given the encoded vector sequence $\mathbf{p}$ and the gloss annotation $\mathbf{g}^{*}$ of the corresponding video, the CTC loss is defined as:
\begin{equation}
    L_{ctc} = -\log p_{ctc}(\mathbf{g}^{*}|\mathbf{p}),
\end{equation}
where $p_{ctc}$ is a probability to generate the given gloss sequence with the condition of the given encoded vector sequence. 

There are many different potential paths to align the encoded vectors with the given gloss sequence. 
Denote $\mathcal{M}^{-1}(\mathbf{g}^{*})$ as the set of all these paths. For a particular path $\mathbf{z} = \{\mathbf{z}_{0},...,\mathbf{z}_{T}\} \in \mathcal{M}^{-1}(\mathbf{g}^{*})$, the probability to obtain this path is in line with the probability computed in $\mathbf{Y}^{g}$: 
\begin{equation}
    p(\mathbf{z}|\mathbf{p})=\prod_{t}y^{g}_{z_{t}t}.
\end{equation}
Hence, the probability for all potential paths, which is exactly the probability $p_{ctc}$, can be computed as:
\begin{equation}
    p_{ctc}(\mathbf{g}|\mathbf{p})=\sum_{\mathbf{z}\in \mathcal{M}^{-1}(\mathbf{P})}p(\mathbf{z}|\mathbf{p}).
\end{equation}

For the generated words in a spoken language, the alignment between the words and the glosses is often not required due to the lack of proper orders. Therefore, a general cross-entropy loss is used to measure the error of each word in a sequence. 
In detail, the loss can be written as: 
\begin{equation}
    L_{ce} = -\log \prod_{l} p(\mathbf{w}_{l}|\hat{\mathbf{w}}_{0},...,\hat{\mathbf{w}}_{l-1},\hat{\mathbf{g}}_{0},...,\hat{\mathbf{g}}_{T}).
\end{equation}
Note that the softmax function to compute the probability is embedded in the computations of the output of the LSTM in the stage 2 decoder. 

Therefore, the total loss is a linear combination of the two loss functions $L_{ctc}$ and $L_{ce}$ with an additional regularization term for the parameters $\Theta$ of the proposed architecture,
\begin{equation}
    L = \lambda_{ctc}L_{ctc} + \lambda_{ce}L_{ce} + \lambda_{r}||\Theta||,
\end{equation}
where $\lambda_{ctc}$, $\lambda_{ce}$ and $\lambda_{r}$ are the weights associated with the three losses and can be tuned as hyper-parameters during the optimization. 

\section{Experimental Results}
\label{sec:exp}

\subsection{Datasets \& Evaluation Metrics}
The proposed method was evaluated on two widely used benchmark sources: PHOENIX-2014, PHOENIX-2014-T~\cite{Camgoz2017} and Chinese Sign Language Recognition (CSL)~\cite{huang2018video,pu2019iterative, zhou2019dynamic}.
PHOENIX-2014 contains videos from PHOENIX TV station, which includes the weather forecast content featured with signers over a period of three years. 
The videos were collected with a resolution of 210 by 260 at 25 frames per second (fps) using a stationary color camera.
The dataset was annotated with sign language glosses and texts in German spoken language. The vocabulary size is 1,115 for sign glosses and 3,000 for German.
The CSL dataset contains two subsets: Split I and Split II. In this study, Split II was adopted, which contains 100 sign language sentences related to the daily life with a vocabulary size of 178. Each sentence in the split was performed by 50 signers each of whom repeated the signing for 5 times. These sentences were recorded in RGB videos with a spatial resolution 1280 by 720 at 30 fps. 
Among the 100 sentences, 94 sentences were in the training set and 6 sentences were in the test set. 

In terms of the evaluation metrics to evaluate the performance, 
two metrics are adopted: word error rate (WER) and bilingual evaluation understudy (BLEU) score, which are widely used for natural language processing (NLP)~\cite{papineni2002bleu}. 
WER measures the recognition performance at the gloss level, whilst BLEU scores measure the performance of the translation. 
BLEU was first proposed to measure the performance of machine translation by comparing the recall and the precision of n-grams.

\subsection{Implementation Details}

To construct input graphs, human skeletons were extracted by two algorithms, HRNet \cite{wang2020deep} and OpenPose~\cite{8765346}. In detail, the coordinates can be extracted for the key points and the key body regions that are used in this study. For high-level graphs, key body regions of the face, the left hand and the right hand were extracted using a window of size 24 by 24 of which the center was located on the corresponding detected key point. Appearance and motion features for these regions were further computed to represent the vertices by using ResNet-152~\cite{he2016deep} and TVL1-flow~\cite{zach2007duality}, respectively. The dimension of these feature vectors is 1,024. 
For fine-level graphs, 29 landmarks and 21 joints were obtained for face and hands, respectively. The coordinates detected by the skeleton detection algorithms were adopted as vertex-level features directly.

Our proposed method was implemented with PyTorch. An Adam optimizer with an initial leaning rate 0.001, and 30 epochs were used to train the model. 
At the validation stage, hyper-parameters $\lambda_{ctc}$ and $\lambda_{ce}$ were set to 0.5 and 0.5, respectively; the temporal window size to construct the spatio-temporal graphs was set to 3 and further discussions are provided in Section~4.5.

\begin{figure}
    \centering
    \includegraphics[width=0.4\textwidth]{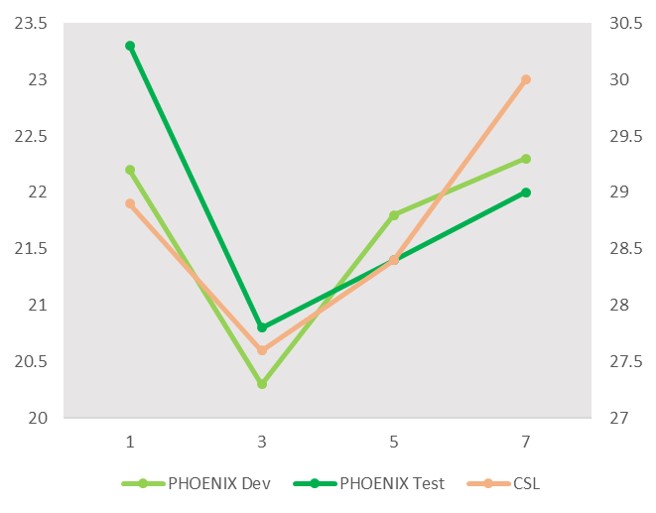}
     \caption{Effect of the windows size on recognition performance (lower is better). WER scores on PHONIX (left y-axis) WER scores on CSL (right y-axis) vs. window size (x-axis).}
     \label{fig:window_size}
\end{figure}

\begin{table}[hbtp]
\centering
\caption{Recognition Performance on PHOENIX-2014. Metric: WER: Lower is better.}
\label{tab:comparison_wer}

\begin{tabular}{p{3.5cm}|p{0.9cm}p{0.9cm}p{0.9cm}}
\hline
Dataset                          & \multicolumn{2}{c}{Phoenix}  & CSL \\ \hline
Subset                           & Test           & Dev           & -     \\ \hline
IAN \cite{pu2019iterative}       & 36.7           & 37.1          & 32.7  \\ 
DenseTCN \cite{Guo2019}          & 36.5           & 35.9          & 44.7      \\
CNN-LSTM-HMM \cite{Koller2019}   & 26.0           & 26.0          & - \\
DNF \cite{cui2019deep}           & 24.4           & 23.8          & -   \\
STMC \cite{Zhou2020}             & 20.7           & 21.1          & 28.6 \\
HLSTM \cite{guo2018hierarchical} & -              & -             & 48.7   \\ \hline
Ours                             & \textbf{19.8}  & \textbf{19.5} & \textbf{27.6} \\ \hline
\end{tabular}
\end{table}

\begin{table*}[hbtp]
\centering
\caption{Translation Performance on PHOENIX-2014-T (Phoenix) and CSL. BLEU: Higher is better.}
\label{tab:comparison}
\begin{tabular}{l|p{0.85cm}p{0.9cm}p{0.8cm}|p{0.9cm}p{0.9cm}p{0.8cm}|p{0.9cm}p{0.9cm}p{0.8cm}|p{0.9cm}p{0.9cm}p{0.8cm}}
\hline
Metric                           & \multicolumn{3}{c|}{BLEU-1}                    & \multicolumn{3}{c|}{BLEU-2}                    & \multicolumn{3}{c|}{BLEU-3}                    & \multicolumn{3}{c}{BLEU-4}                    \\ \hline

Dataset                       & \multicolumn{2}{c}{Phoenix}                  & CSL           & \multicolumn{2}{c}{Phoenix}                 & CSL           & \multicolumn{2}{c}{Phoenix}                   & CSL           & \multicolumn{2}{c}{Phoenix}                   & CSL           \\ \hline
Subset                           &  Test           & Dev           & -    & Test           & Dev           & -   & Test           & Dev           & -    & Test           & Dev           & -      \\ \hline
NSLT \cite{Camgoz2018}                        & 43.3          & 42.9         &     -          & 30.4          & 30.3         &     -          & 22.8          & 22.02         &       -        & 18.1          & 18.4         &   -            \\
SLTT\cite{yin2020sign}& 44.95         & 48.27         &     -          & 36.53          & 35.20         &     -          & 29.30          & 27.47         &      -         & 24.00        & 22.47         &     -          \\ 
JSL\cite{camgoz2020sign} & 46.61& 47.26& - & 33.73& 34.40&- & 26.19 & 27.05&- & 21.32 & 22.38 & - \\
Ours            & \textbf{45.2}  & \textbf{46.1} & \textbf{49.1} & \textbf{34.7}  & \textbf{33.4} & \textbf{33.1} & \textbf{27.1}  & \textbf{27.5} & \textbf{22.7} & \textbf{22.3}  & \textbf{22.6} & \textbf{17.8} \\ \hline
\end{tabular}
\end{table*}

\begin{table*}[htbp]
    \centering
      \caption{Ablation studies on PHOENIX-2014-T (Phoenix) and CSL. WER: Lower is better, BLEU: Higher is better. \checkmark { } (or \xmark) indicates the inclusion (or exclusion) of a specific mechanism (S: spatial graph, T: temporal graph, H: hierarchical graph).}
\begin{tabular}{c|p{0.5cm}p{0.5cm}p{0.5cm}@{\,}|@{\ }p{0.5cm}p{0.5cm}p{0.5cm}|@{\ }p{0.5cm}p{0.5cm}p{0.5cm}|@{\ }p{0.5cm}p{0.5cm}p{0.5cm}|@{\ }p{0.5cm}p{0.5cm}p{0.5cm}|@{\ }p{0.5cm}p{0.5cm}p{0.5cm}}
\hline
\multirow{3}{*}{Model}                       & \multicolumn{3}{c|@{\ }}{Method}           & \multicolumn{3}{c|@{\ }}{WER}                      & \multicolumn{3}{c|@{\ }}{BLEU-1}                   & \multicolumn{3}{c|@{\ }}{BLEU-2}                   & \multicolumn{3}{c|@{\ }}{BLEU-3}                   & \multicolumn{3}{c}{BLEU-4}                   \\ \cline{2-19}
\multicolumn{1}{c|}{}    & \multicolumn{3}{c|@{\ }}{Dataset}         & \multicolumn{2}{c}{Phoenix}  & CSL           & \multicolumn{2}{c}{Phoenix}  & CSL           & \multicolumn{2}{c}{Phoenix}  & CSL           & \multicolumn{2}{c}{Phoenix}  & CSL           & \multicolumn{2}{c}{Phoenix}  & CSL           \\ \cline{2-19}
\multicolumn{1}{c|}{}    & S          & T          & H          & Test          & Dev           & -             & Test          & Dev           & -             & Test          & Dev           & -             & Test          & Dev           & -             & Test          & Dev           & -             \\ \hline
\multicolumn{1}{c|}{I}   & \xmark     & \xmark     & \xmark     & 35.8          & 35.4          & 32.1          & 43.2          & 43.3          & 48.1          & 30.2          & 30.1          & 28.1          & 22.1          & 21.9          & 16.6          & 18.0          & 18.2          & 14.3          \\
\multicolumn{1}{c|}{II}  & \xmark     & \checkmark & \xmark     & 23.4          & 23.2          & 30.8          & 43.4          & 43.4          & 48.4          & 30.9          & 30.7          & 29.1          & 23.2          & 23.1          & 17.1          & 18.5          & 18.8          & 14.6          \\
\multicolumn{1}{c|}{III} & \checkmark & \xmark     & \xmark     & 22.4          & 22.5          & 29.3          & 43.6          & 43.5          & 48.8          & 31.7          & 31.2          & 31.3          & 24.7          & 25.2          & 19.9          & 19.2          & 19.7          & 15.9          \\
\multicolumn{1}{c|}{IV}  & \checkmark & \checkmark & \xmark     & 20.7          & 20.3          & 28.6          & 43.7          & 43.8          & 49.1          & 32.1          & 32.5          & 32.1          & 25.9          & 26.1          & 21.1          & 20.8          & 21.3          & 17.4          \\
\multicolumn{1}{c|}{V}   & \checkmark & \checkmark & \checkmark & \textbf{19.8} & \textbf{19.5} & \textbf{27.6} & \textbf{45.2} & \textbf{46.1} & \textbf{49.1} & \textbf{34.7} & \textbf{33.4} & \textbf{33.1} & \textbf{27.1} & \textbf{27.5} & \textbf{22.7} & \textbf{22.3} & \textbf{22.6} & \textbf{17.8} \\ \hline
\end{tabular}
    \label{tab:ablation1}
\end{table*}

\begin{table}[htbp]
\caption{Ablation Study on PHOENIX-2014T: Different Features.}
\label{tab:ablation2}
\begin{tabular}{l|c|c|c|c|c}
\hline
\multicolumn{6}{c}{Dev}                 \\ \hline
 HST-GNN              & WER  & B1   & B2   & B3   & B4     \\ \hline
w/o appearance & 23.8 & 44.1 & 31.3 & 26.2 & 20.7 \\ \hline
w/o motion     & 20.1 & 42.3 & 32.2 & 24.9 & 19.1 \\ \hline
w/o pose       & 22.3 & 43.7 & 32.8 & 25.1 & 20.3 \\ \hline
\multicolumn{6}{c}{Test} \\ \hline
 HST-GNN & WER  & B1   & B2   & B3   & B4 \\ \hline
w/o appearance &  23.7 & 44.8 & 31.8 & 26.7 & 20.1 \\ \hline
w/o motion     &  20.3 & 42.1 & 32.9 & 24.5 & 17.9 \\ \hline
w/o pose       &  21.9 & 40.1 & 33.1 & 23.9 & 19.3 \\ \hline
\end{tabular}
\end{table}

\subsection{Overall Performance}

To demonstrate the effectiveness of the proposed method, a number of recently proposed methods were compared as shown in Table~\ref{tab:comparison}: iterative alignment network (IAN) \cite{pu2019iterative}, which adopts an iterative alignment network to reduce the gap between videos and generated glosses, enabling a better correspondence between glosses and frames;  
DenseTCN \cite{Guo2019}, which introduced temporal convolutions to efficiently explore temporal patterns; CNN-LSTM-HMM \cite{Koller2019}, which combined LSTM and HMM in language modeling for the construction of gloss sequences and an intermediate synchronization constraints, respectively. 
Deep neural frame (DNF) \cite{cui2019deep}, which incorporates RGB and motion features of body regions to explore fine-level sign language patterns; 
Spatial-temporal multi-cue network (STMC) \cite{Zhou2020}, which involved a dense network to characterize the spatial and the temporal dependencies to improve the recognition quality; 
Hierarchical LSTM (HLSTM) \cite{guo2018hierarchical}, which was proposed to extract multiple levels of attention with adaptive online key clip mining; 
Neural sign language translation (NSLT) \cite{Camgoz2018}, which was the first study using CNN-LSTM sign language translation in an end-to-end manner; 
Neural language translation with transformer (SLTT) \cite{yin2020sign}, which utilized transformers for sign language translation based on STMC \cite{Zhou2020}. 


It can be observed that the methods with fine-level regional body patterns (e.g., DNF and STMC) achieved better performance compared to IAN. The introduction of the relationships between these body regions further improved the performance as STMC was better than DNF. The explicit inclusion of the temporal clues could also help to improve the SLR performance, for example, CNN-LSTM-HMM was superior to IAN. 
Moreover, the adoption of the transformer helped to increase the recognition performance, such as NSLT vs SLTT, which suggested that proper transformer designs could be beneficial for the sign language modelling. 
In terms of WER, our method achieved 19.8 on the Phoenix-2014-T dev set, 19.5 on Phoenix-2014-T test set and 27.6 on the CSL dataset; for BLEU-1 score, our method achieved 45.2 on the Phoenix-2014-T dev set, 46.1 on Phoenix-2014-T test set and 49.1 on the CSL dataset. This indicates that our method achieved the state-of-the-art performance on the two benchmark datasets for sign language recognition and translation.

\subsection{Effect of Temporal Window Size}
It is anticipated that the temporal window size impacts the performance of our proposed HST-GNN as a larger temporal window implies that longer time dependencies can be captured. 
Nonetheless, a long temporal perception could increase the model complexity and introduce unnecessary historical information. Therefore, 
the trade-off to select a proper window size was investigated. The results are shown in~\autoref{fig:window_size} in terms of the WER and BLEU scores on the two benchmarks. 
It can be observed that the recognition performance increased when the window size was changed from 1 (i.e., only the current state without any historical patterns) to 3. However, if the window size further increased, the performance was negatively impacted for the two datasets. Hence, a window size of 3 was used as a proper choice in line with the experimental results.

\subsection{Ablation Study}
To further explore how the proposed mechanisms work for SLR under different configurations, ablation studies were conducted for the spatial, temporal and hierarchical modules in HST-GNN. A baseline model denoted as Model I was introduced without involving any graph patterns. Next, a number of architectures involving parts of these mechanisms were investigated on top of the baseline model: Model II involves temporal graphs; Model III involves spatial graphs; Model IV involves both spatial and temporal graphs; and Model V considers spatial, temporal, and hierarchical graphs all, which is the full version of the proposed HST-GNN. 
The results are shown in \autoref{tab:ablation1} and \autoref{tab:ablation2}. It can be observed that using the spatial or the temporal graph patterns individually improved the performance compared to the baseline model in terms of WER and BLEU scores on both benchmark datasets. 
Introducing both the spatial and temporal graph patterns clearly enhanced the performance further. 
Lastly, hierarchical modelling also showed its effectiveness for SLR.

\subsection{Qualitative Analysis}

\begin{figure}[htbp]
    \centering    
    \includegraphics[width=0.45\textwidth]{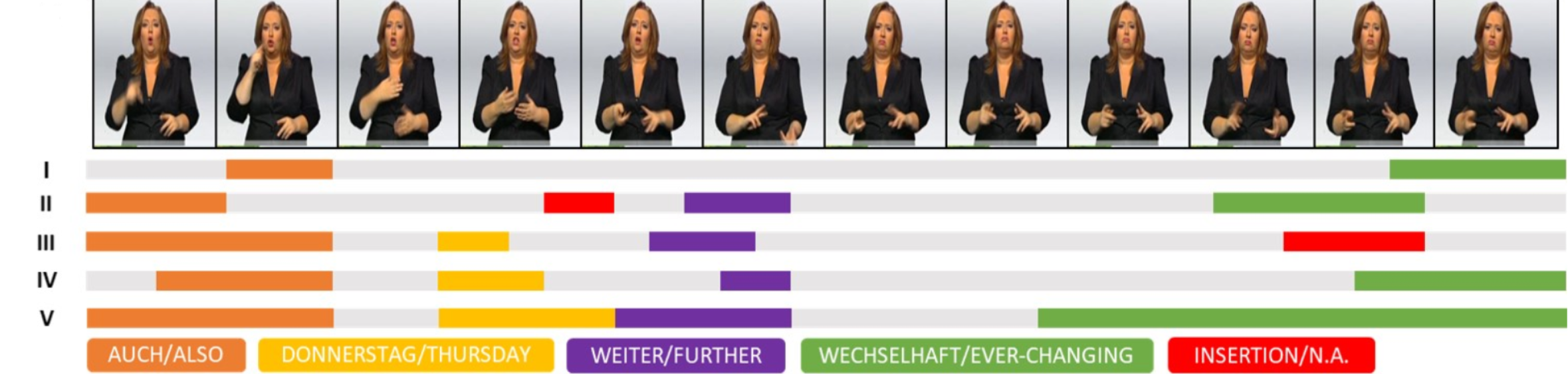}
    \caption{Sample results of our proposed HST-GNN model under different configurations. 
    CTC objectives are illustrated to indicate the associations of the gloss (with English translation) and the video frames. \textit{INSERTION} means a wrong gloss prediction outside the ground truth gloss annotation. }
    \label{fig:aligments}
\end{figure}

To understand the proposed methods, two examples, which were featured by two different signers, were illustrated in Figure~\ref{fig:aligments}. Video frames, the frame-level gloss predictions of the five models and the ground truth of the gloss sequence are presented. From the first example, it can be observed that the \textit{Baseline} method (Model I) missed the glosses DONNERSTAG and WEITER. With the temporal graphs (Model II), the gloss WEITER can be detected, whilst the gloss DONNERSTAG was still missed and a new insertion error occurred. Similarly, for the case only with the spatial graphs (Model III), three glosses were detected and one was missed with an insertion error. By introducing the spatial and the temporal mechanisms simultaneously (Model IV), all glosses were detected correctly without insertion errors. The hierarchical mechanism further improved the gloss predictions, by which the number of frames with glosses were increased significantly. The present of both graph can successfully detect the all of the glosses.



\section{Conclusion}
\label{sec:conclusion}
In this paper, a novel neural network, namely HST-GNN, is presented for sign language understanding. Hierarchical graphs are introduced to characterize visual signing content, which include high-level graphs and fine-level graphs associated with key body regions. HST-GNN follows an encoder-decoder framework: the encoder adopts graph convolutions and graph transformers with an adjacency matrix based connection to explore both the global and local graph properties; the decoder reconstructs the glosses and the spoken language script in line with the latent embedding. Experiments on two widely used dataset including PHOENIX-2014-T and CSL were conducted and the results clearly demonstrated the effectiveness of the proposed HST-GNN. In our future work, we will investigate additional graph properties to improve the performance of SLT.

{\small
\bibliographystyle{ieee_fullname}
\bibliography{egbib}
}

\end{document}